# Meta-Embeddings for Natural Language Inference and Semantic Similarity tasks

Shree Charran R, Rahul Kumar Dubey, *Senior Member IEEE*

**Abstract**— Word Representations form the core component for almost all advanced Natural Language Processing (NLP) applications such as text mining, question-answering, and text summarization, etc. Over the last two decades, immense research is conducted to come up with one single model to solve all major NLP tasks. The major problem currently is that there are a plethora of choices for different NLP tasks. Thus for NLP practitioners, the task of choosing the right model to be used itself becomes a challenge. Thus combining multiple pre-trained word embeddings and forming meta embeddings has become a viable approach to improve tackle NLP tasks. Meta embedding learning is a process of producing a single word embedding from a given set of pre-trained input word embeddings. In this paper, we propose to use Meta Embedding derived from few State-of-the-Art (SOTA) models to efficiently tackle mainstream NLP tasks like classification, semantic relatedness, and text similarity. We have compared both ensemble and dynamic variants to identify an efficient approach. The results obtained show that even the best State-of-the-Art models can be bettered. Thus showing us that meta-embeddings can be used for several NLP tasks by harnessing the power of several individual representations.

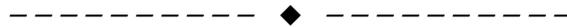

─────────── ◆ ──────────

## 1 INTRODUCTION

Arguably one of the fundamental tasks in natural language processing (NLP) is to represent the meanings of individual words. If one can correctly represent the meanings of individual terms as some representations then we can further use such representations to compose the meanings of broader lexical units, such as sentences, paragraphs, or even the entire document. To tackle this issue, over the last decade the NLP community has proposed numerous different approaches for learning word representations/embeddings. However, all these word embedding are learned using different methods, thus they report varying degrees of success on various tasks. Simply put, when it comes to word embedding, there is no single method proposed so that solves all NLP problems efficiently. The search for better word embedding is never-ending and new methods for learning word embedding are being proposed frequently which capture different aspects of word semantics and industry issues.

However, this plethora of choices can become a bit of a challenge for NLP practitioners. How does one choose the best embedding model out of several available to train for a specific NLP application? Due to time constraints or a sheer number of different word embedding methods available, Trial and error don't seem like a feasible solution.

Thus combining multiple pre-trained word embeddings and forming meta embeddings has become a viable approach to improve word representations. Meta Embedding can use several good individual embeddings and ensemble them so that they cover various aspects of word semantics thus making it even richer and form a larger vocabulary that provides an immense advantage in cross-domain tasks. To be clear, Meta embedding learning is a process of producing a single word embedding from a given set of pre-trained input word embeddings. Two key aspects make meta embedding learning problems different and better from that of learning source embedding.

Firstly, we do not re-train any of the source word embedding learning algorithms during meta embedding. One does not need to assume the availability of the implementations of the methods that were used to produce the source word embeddings. We can directly use pre-trained source word embeddings. This makes the meta embedding learning algorithms independent of the source word embedding learning algorithms, which is great because we can use the same meta embedding learning algorithm to meta embed any given set of word embeddings.

Secondly, we do not assume the availability of the text corpora that were used to train the source word embeddings. This is attractive because sometimes all that you get with source word embeddings is the pre-trained vectors and not the language resources such as text corpora or dictionaries that were used to train

────────────

• *Shree Charran R is with the Department of Management Studies, Indian Institute of Science, Bengaluru 560076, and India (E-mail: shreer@iisc.ac.in)*
• *Rahul Kumar Dubey is with Robert Bosch Engineering and Business Solutions Private Limited, Bengaluru, Karnataka 560095, India (E-mail:rahul.dubey2011@ieee.org)*

those source word embeddings. This could be because, for example, the language resources are proprietary and the authors of the source embedding learning algorithms are not allowed to release the language resources they used. It also makes the meta embedding learning algorithms attractive because they do not require language resources during training. Moreover, even if you had access to the language resources, it might be computationally costly to re-train the source word embedding learning methods on those language resources by yourself, especially if you do not have access to the hardware such as GPU units. In this paper, we evaluate the performance of Ensemble and Dynamic Meta-embeddings on Natural Language Inference and Semantic Similarity tasks.

## 2 RELATED WORK

### 2.1 SENTENCE EMBEDDINGS:

Sentence embeddings is a single representation for a sentence rather than a word. The first major approach is Word embedding average sentence encoders where it takes the weighted average of word embeddings. The major drawback of this approach is that word order and polysemy are compromised, Hence particularly ineffective in sentiment analysis and but effective in sentence similarity tasks[23]. The second major approach is Complex contextualized sentence encoders, such as Long Short Term Memory Networks (LSTM) [24] or Transformers [25] Contextualized encoders can be pre-trained as unsupervised language models [26], but they are usually improved on supervised transfer tasks such as Natural Language Inference [21].

### 2.2 SENTENCE EMBEDDINGS EVALUATION:

With continuous developments of word and sentence representations, there is a need for a standardized evaluation of representations. SentEval[12], is an open-sourced toolkit available for evaluating the quality of universal sentence representations. [12] consists of a variety of tasks, including binary and multi-class classification, Natural Language Inference (NLI), etc. This work aimed to provide a more centralized way to evaluate sentence representations. In [13], the authors used the evaluation tasks suggested by [12] to probe the quality of major sentence representations to assess the state-of-the-art performers in each category. ELMo [3], USE [4] & Infersent [5] came out as the leaders in all ten classes of assessment.

### 2.3 META-EMBEDDINGS:

The performance of the word embedding depends on different factors such as the size and genre of training corpora as well as the type of training methodology used. This led to the development of meta-embedding approaches from ensemble from pre-trained embeddings[7][9][10][11].

Diverse approaches to learning meta-embedding have been proposed and can be broadly classified into two categories: (a) basic linear methods such as averaging or concatenation, or low-dimensional projection; [7] (b) non-linear framework [9][10][11] which use auto-encoding or transformation to learn meta-embedding as a shared representation between each and a common representation

In [7], and effective a meta-embedding learning framework namely 1TON was proposed. 1TON projects a meta-embedding of a word into the source embeddings using separate projection. These projection matrices were learned by minimizing the sum of squared Euclidean distance between the projected source embeddings and the corresponding source embeddings for all the words in the vocabulary. The authors also proposed an extension (1TON+) to their meta-embedding learning framework that first predicts the source word embeddings for out-of-vocabulary words in a particular source embedding, using the known word embeddings. Next, the 1TON method is applied to learn the meta-embeddings for the union of the vocabularies covered by all of the source embeddings. Experimental results in semantic similarity prediction, word analogy detection, and cross-domain POS tagging tasks show the effectiveness of both 1TON and 1TON+.

In [8], they proposed a geometric framework for learning meta-embedding of words from various sources of word embeddings. The geometric averaging and geometric concatenation outperformed the plain averaging and the plain concatenation models.

In [9], the authors extended their work on autoencoders and developed three variants i.e. Decoupled Autoencoded, Concatenated Autoencoded & Average Autoencoded. Of these, the Average Autoencoded outperformed the other two and baseline models

## 3 EMBEDDING METHODS

In this section we explore three ensemble methods: Concatenation, Singular Value Decomposition, Generalized Canonical Correlation Analysis, and two Dynamic methods: Dynamic meta-embeddings and Contextualized dynamic meta-embeddings.

### 3.1 Concatenation (CON):

In Concatenation, the meta-embedding M is the concatenation of all the individual embeddings $E_i$ of dimension $|E_i|$ (i=1,2…n) or concatenation of select individual embeddings. The main intuition behind Concatenation is that: by combining different embedding sets, we can capture all of the information in the individual embeddings.

$$M_j^{Con} = [E_{1,j}, E_{2,j}, \ldots, E_{n,j}] \quad (1)$$

The dimensionality of CON meta-embeddings $M_j^{Con}$ is k. where,

$$k = \sum_1^n |Ei| \quad (2)$$

The major drawback we need to keep in mind while concatenating embeddings it becomes inefficient as we combine more and more embeddings.

## 3.2 SINGULAR VALUE DECOMPOSITION (SVD):

The major disadvantage of concatenation as a method for creating meta-embeddings is the curse of dimensionality [1] proposed the use of SVD to reduce the dimensionality of the meta-embeddings created from concatenation. Given a set of CON representations for n-words, each of dimensionality k, we compute an SVD decomposition $C = USV^T$ of the corresponding nxk matrix C. We then use $U_d$, the first d dimensions of U, as the SVD meta-embeddings of the n-words. L2-normalization is performed to embeddings; similarities of SVD vectors are computed as dot products. SVD is guaranteed to produce the best (in the sense of least square error) rank k approximation of a matrix.

## 3.3 Generalized Canonical Correlation Analysis (GCCA):

Given three or more random vectors x1,x2,x3; GCCA finds linear projections such that $\theta_1^T x_1, \theta_2^T x_2, \theta_3^T x_3$ are maximally correlated. [14] showed a variant of GCCA can be reduced to a generalized eigenvalue problem on block matrices

$$\rho \begin{pmatrix} \sum_{11} & 0 & 0 \\ 0 & \sum_{\ldots} & 0 \\ 0 & 0 & \sum_j \end{pmatrix} \begin{bmatrix} \theta_1 \\ \ldots \\ \theta_j \end{bmatrix} \quad (3)$$

$$= \rho \begin{pmatrix} 0 & \sum_{\ldots} & \sum_{1,j} \\ \sum_{\ldots} & 0 & \sum_{\ldots} \\ \sum_{j,1} & \sum_{\ldots} & 0 \end{pmatrix} \begin{bmatrix} \theta_1 \\ \ldots \\ \theta_j \end{bmatrix} \quad (4)$$

where,

$$\sum_{j,j'} = E_{s \in S}\left[(M_j(s) - \mu_j)(M_{j'}(s) - \mu_{j'})^T\right] \quad (5)$$

$$\mu_j = E_{s \in S}[(M_j(s))] \quad (6)$$

For stability, we add $\frac{\tau}{d_j} \sum_{n=1}^{d_j} diag(\sum_{j,j\cdot})_n$ to $diag(\sum_{j,j})$. Where $\tau$ is a hyperparameter. We stack the eigenvectors of the top-d eigenvalues into matrices. We define the GCCA meta-embedding of sentence s' as:

$$M^{gcca}(s') = \sum_{j=1}^{J} \odot (M_j(s') - \mu_j) \quad (7)$$

As suggested by [26] The value $\tau$ is set as 10 for best results.

## 3.4 DYNAMIC META-EMBEDDINGS (DME):

In DME, the embeddings are projected onto a common d'-dimensional space using learned linear functions

$M_{i,j}' = P_i w_{i,j} + b_i$ (i = 1,2.., n) where $P_i \in \mathbb{R}^{d' \times d_i}$ and $b_i \in \mathbb{R}^{d'}$. The combination of the weighted sum of projected embeddings give

$$M_j^{DME} = \sum_{i=1}^{n} \alpha_{i,j} M'_{i,j} \quad (8)$$

where, $\alpha_{i,j} = g(\{M'_{i,j}\}_{j=1}^{s})$ are the scalar weights from a self-attention mechanism:

$$\alpha_{i,j} = g(M'_{i,j}) = \theta(a. M'_{i,j} + b) \quad (9)$$

where $a \in \mathbb{R}^{d'}$ and $b \in \mathbb{R}^{d'}$ are the learned parameters and $\theta$ is a softmax.

DME uses a mechanism that is dependent only on the word projections themselves. Each word projection gets multiplied by a learned vector a which results in a scalar — so for N different projections (which corresponds N different embedding sets), we will have N scalars. These scalars are then passed through the softmax function and the results are the attention coefficients. Then these coefficients are used to create the meta-embedding, which is the weighted sum (weighted using the coefficient) of the projections. Figure 1 helps visualize DME operations.

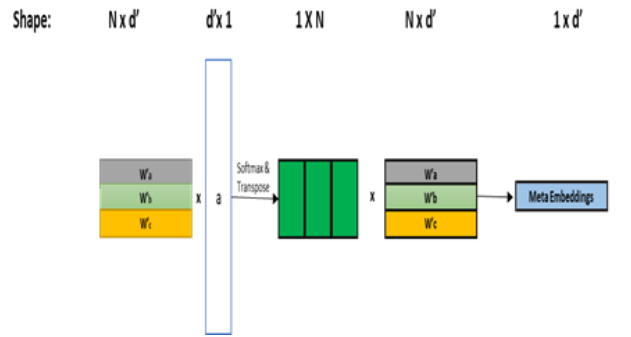

FIGURE 1: DME EMBEDDING

## 3.5 CONTEXTUALIZED DYNAMIC META-EMBEDDINGS (DME):

CDME, adds context into the mix, using a Sentence encoder. The difference from DME is in way attention coefficients are calculated. Just like DME, first, the sequences get projected, and then passed through the Sentence encoder. Then, instead of using the word itself, the concatenated hidden state of the forward and backward LSTMs (Sentence encoder) (in the word's corresponding index) and the vector are used to calculate the attention coefficient.

$$\alpha_{i,j} = g(M'_{i,j}) = \theta(a. h_j + b) \quad (10)$$

where $h_j \in \mathbb{R}^{2m}$ is the j$^{th}$ hidden state of a BiLSTM taking $(M'_{i,j})$ as input, $a \in \mathbb{R}^{2m}$ and $b \in \mathbb{R}^m$. In [27] authors suggest we take m = 2, which makes the contextualization very efficient.

Sentence encoder : a standard bidirectional LSTM encoder with max-pooling (BiLSTM-Max), is used:

$$\overrightarrow{h_j} = \overrightarrow{LSTM}_j(w_1, w_2, \ldots, w_j) \quad (11)$$

$$\overleftarrow{h_j} = \overleftarrow{LSTM}_j(w_j, w_{j+1}, \ldots, w_s) \quad (12)$$

The hidden states are concatenated at each timestep to obtain the final hidden states, after which a max-pooling operation is applied over their components to get the final sentence representation:

$$h = \max\{(\overrightarrow{h_j}, \overleftarrow{h_j})\}_{j=1,2\ldots n} \quad (13)$$

Figure 2 helps visualize the CDME operations.

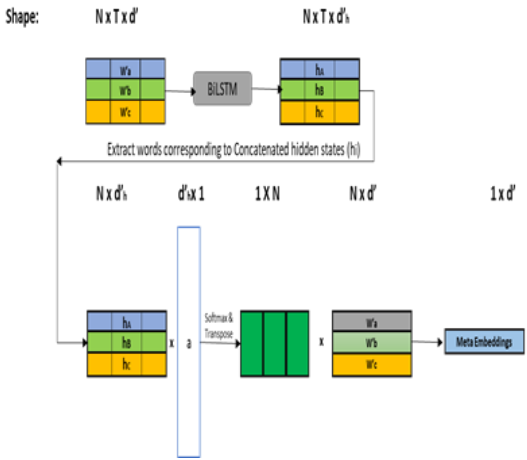

FIGURE 2: CDME EMBEDDING

## 4 EXPERIMENTAL SETUP

### 4.1 SOURCE EMBEDDINGS:

In this section, we explain in brief about the selected pre-trained models which we will use as source models:

#### 4.1.1 EMBEDDINGS FROM LANGUAGE MODELS (ELMO)[3]:

ELMo is a deep contextualized representation of words that models both (1) complex word use characteristics like syntax and semantic structure and (2) polysemy modeling. Such word vectors are learned functions of a deep bidirectional language model 's internal states (biLM), which is pre-trained on a large text corpus. ELMo can easily be applied to existing models and greatly improve the state of the art across a wide spectrum of difficult NLP issues, including answering questions, classification textual entailment, etc. We shall use the ELMo (BoW, all layers, 5.5B) embedding whose model was trained on a dataset with 5.5B tokens consisting of Wikipedia (1.9B) and WMT (3.6B).

#### 4.1.2 UNIVERSAL SENTENCE ENCODER (USE)[4] :

USE encodes text into vectors of large dimensions that can be used for semantic similarity, clustering, and other natural language tasks. Both have accuracy and computational need trade-off. Although the one with a Transformer encoder has higher accuracy, it is more expensive in computational terms. The one with DNA encoding is less costly computationally and with comparatively lesser accuracy.

TABLE 1
KEY POINTS OF THE SOURCE EMBEDDINGS

| Model | Training method | Embedding size |
|---|---|---|
| ELMo (BoW, all layers, 5.5B) | Self-supervised | 3072 |
| InferSent (AllNLI) | Supervised | 4096 |
| USE (Transformer) | Supervised | 512 |

#### 4.1.3 INFERSENT[5]:

InferSent incorporates supervised training to obtain sentence embeddings. The sentence encoders are trained using the Stanford Natural Language Inference (SNLI) dataset, which consists of 570k human-generated English sentence-pairs and it is considered one of the largest high-quality labeled datasets for building sentence semantics understanding. The authors found the best results were achieved with a bi-directional LSTM (BiLSTM) encoder.

In [6] the authors observed ELMo achieved the best results in 5 out of 9 tasks downstream tasks. While InferSent achieved the best results in the paraphrase detection as well as in the SICK-E (entailment)., the Universal Sentence Encoder (USE) using the Transformer model achieved excellent results in the semantic relatedness and textual similarity tasks. Thus we consider these as the SOTA to use as source embeddings. Table 1 gives the key points of outsourcing embeddings

## 4.2 EVALUATION TASKS:

In this section, we explain the Natural Language Inference and Semantic Similarity tasks employed to evaluate the performance of the Meta-Embeddings.

### 4.2.1 SENTENCES INVOLVING COMPOSITIONAL KNOWLEDGE SEMANTIC ENTAILMENT/ RELATEDNESS (SICK_E & SICK_R)[15]:

It is a crowdsourced dataset consisting of 10,000 English sentence pairs. Each sentence pair in the SICK dataset is annotated to mark the following two aspects (i) the degree relatedness between a sentence pair (on a scale of 1-5), and (ii) whether one sentence entails or contradicts or is neutral the other (considering both directions). We learn to predict the probability distribution of relatedness scores

### 4.2.2 SEMANTIC TEXTUAL SIMILARITY BENCHMARK (STS-B)[16-20]:

These datasets consist of sentence pair taken from news snippets, headlines, news conversations, headlines, image descriptions, etc. and is labeled with a similarity score between 0 and 5. The task is to evaluate how the cosine distance (similarity) between two sentences correlate with a human-labeled similarity score by using Pearson correlations. We use STS tasks from 2012 to 2016.

### 4.2.3 STANFORD NATURAL LANGUAGE INFERENCE (SNLI)[21]:

The SNLI dataset [21] is a collection of 570k human-written English supporting the task of natural language inference (NLI), also known as recognizing textual entailment (RTE) which consists of predicting whether two input sentences are entailed, neutral or contradictory. SNLI was specifically designed to serve as a benchmark for evaluating text representation learning methods.

### 4.2.4 PARAPHRASE DETECTION THE MICROSOFT RESEARCH PARAPHRASE CORPUS (MRPC)[22]:

This data - set consists of pairs of sentences that were extracted from Web news sources. The Sentence pairs are human-annotated depending on how they can capture the relation of paraphrase / semantic equivalence. It only has 2 classes, i.e. the purpose is to predict whether or not the sentences are paraphrases. Table 2 gives a sample of all the datasets used for evaluation.

TABLE 2
Samples from all evaluation dataset

| Name | Size | Sentence 1 | Sentence 2 | Score |
|---|---|---|---|---|
| SICK-R | 10000 | "A girl is playing the guitar" | "A girl opens a package that contains headphones" | 1.6 |
| SICK-E | 10000 | "A man is sitting on a chair and rubbing his eyes" | "There is no man sitting on a chair and rubbing his eyes" | Contradiction |
| STS-B | 8700 | Liquid ammonia leak kills 15 in Shanghai | Liquid ammonia leak kills at least 15 in Shanghai | 4.6 |
| SNLI | 570000 | "A small girl wearing a pink jacket is riding on a carousel." | "The carousel is moving." | entailment |
| MRPC | 5700 | "The procedure is generally performed in the second or third trimester" | "The technique is used during the second and, occasionally, third trimester of pregnancy." | paraphrase |

## 4.3 EVALUATION METRICS:

### 4.3.1 COSINE SIMILARITY

Cosine similarity is a metric used to determine how similar two embeddings (text). It is given as the dot product of two vectors(A, B). Its denoted as :

$$\text{Similarity}(A,B) = \frac{A.B}{\|A\| * \|B\|} \quad (14)$$

$$= \frac{\sum_{i=1}^{n} A_i * B_i}{\sqrt{\sum_{i=1}^{n} A_i^2 * \sum_{i=1}^{n} B_i^2}} \quad (15)$$

Where:

A, B are two token/sentence embedding in vector form.

The Cosine similarity produces a score between 0-1. Where 1.00 indicates the maximum similarity the 0.00 the least similarity. The SICK_R & STS-B dataset is scored between 0-5. Thus we scale the cosine similarity score accordingly.

### 4.3.2 PEARSON'S CORRELATION COEFFICIENT ($r_{xy}$):

Pearson's correlation coefficient measures the strength of the association between the two continuous variables. In our case, it measures the strength of the association between the cosine similarity scores of the original dataset and the prediction.

It is given by the formula :

$$r_{xy} = \frac{\sum_{i=1}^{n} X_i Y_i - n\bar{X}\bar{Y}}{\sqrt{\sum_{i=1}^{n}(X_i - \bar{X})^2 \sum_{i=1}^{n}(Y_i - \bar{Y})^2}} \quad (16)$$

Where:

- $\bar{x}, \bar{y}$ are two sets of the mean of X & Y.
- N is the sample size

### 4.3.3 ACCURACY:

Accuracy is a straight forward metric and is measured as a percentage of total correct.

## 4.4 EXPERIMENTAL DETAILS:

a) All the Evaluations are conducted for individual source embeddings and all combinations of source embeddings.
b) The dimension of concatenated embeddings will be the sum of individual embeddings.
c) The latest pretrained version of the Source embeddings is used and indicated in Table 1.
d) The dimension of SVD & GCCA is set to 3072, the median size of embeddings.
e) The value of $\tau$ (GCCA) as 10
f) tuned on the STS Benchmark development
g) To maintain uniformity the following parameters for Logistic Regression is considered for classification tasks as suggested by [12]:

```
params['classifier'] ={'nhid': 0,
    'optim': 'adam',
    'batch_size': 64,
    'tenacity': 5,
    'epoch_size': 4}
```

h) We have conducted all experiments by SentEval[18] standards.

## 5 RESULTS & DISCUSSION:

We have conducted the evaluation tasks initially for the source models so that we have a benchmark for performance. There are 4 combinations of source embeddings possible (ELMo +USE+ InferSent , ELMo +USE , USE+ InferSent, ELMo + InferSent). These 4 variations of source embeddings are fed to the ensemble models and corresponding results are tabulated.

Table 3: Results of the semantic relatedness and textual similarity tasks
(Pearson correlation coefficient)

| | | STS-12 | STS-13 | STS-14 | STS-15 | STS-16 | STS-B | SICK-R |
|---|---|---|---|---|---|---|---|---|
| Source Embeddings | ELMo | 0.54 | 0.52 | 0.62 | 0.67 | 0.59 | 0.66 | 0.83 |
| | USE | 0.60 | 0.63 | 0.70 | 0.73 | 0.73 | 0.77 | 0.85 |
|  | InferSent | **0.60** | **0.55** | **0.67** | **0.70** | **0.70** | **0.76** | **0.88** |
| CON | All | 0.59 | 0.57 | 0.67 | 0.71 | 0.68 | 0.74 | 0.86 |
|  | - ELMo | 0.61 | 0.60 | 0.70 | 0.73 | 0.73 | 0.78 | 0.88 |
|  | - USE | 0.58 | 0.55 | 0.66 | 0.70 | 0.66 | 0.72 | 0.87 |
|  | - InferSent | 0.60 | 0.61 | 0.69 | 0.73 | 0.69 | 0.75 | 0.88 |
| SVD | All | 0.59 | 0.58 | 0.68 | 0.71 | 0.69 | 0.74 | 0.87 |
|  | - ELMo | 0.62 | 0.61 | 0.71 | 0.74 | 0.74 | 0.79 | 0.89 |
|  | - USE | 0.59 | 0.55 | 0.66 | 0.71 | 0.66 | 0.73 | 0.88 |
|  | - InferSent | 0.60 | 0.60 | 0.68 | 0.74 | 0.70 | 0.76 | 0.89 |
| GCCA | All | 0.63 | 0.61 | 0.72 | 0.76 | 0.71 | 0.81 | 0.92 |
|  | - ELMo | **0.65** | **0.63** | **0.74** | **0.78** | **0.75** | **0.81** | **0.94** |
|  | - USE | 0.63 | 0.61 | 0.72 | 0.76 | 0.73 | 0.79 | 0.92 |
|  | - InferSent | 0.63 | 0.64 | 0.73 | 0.78 | 0.73 | 0.79 | 0.93 |
| DME | All | 0.67 | **0.70** | **0.78** | **0.81** | **0.81** | **0.85** | 0.92 |
|  | - ELMo | 0.65 | 0.69 | 0.76 | 0.79 | 0.79 | 0.84 | 0.90 |
|  | - USE | 0.65 | 0.68 | 0.75 | 0.79 | 0.79 | 0.83 | 0.89 |
|  | - InferSent | 0.63 | 0.67 | 0.74 | 0.77 | 0.77 | 0.81 | 0.87 |
| CDME | All | **0.69** | 0.67 | 0.75 | 0.78 | 0.78 | 0.82 | **0.93** |
|  | - ELMo | 0.65 | 0.69 | 0.76 | 0.79 | 0.79 | 0.84 | 0.90 |
|  | - USE | 0.65 | 0.68 | 0.75 | 0.79 | 0.79 | 0.83 | 0.89 |
|  | - InferSent | 0.63 | 0.67 | 0.74 | 0.77 | 0.77 | 0.81 | 0.87 |

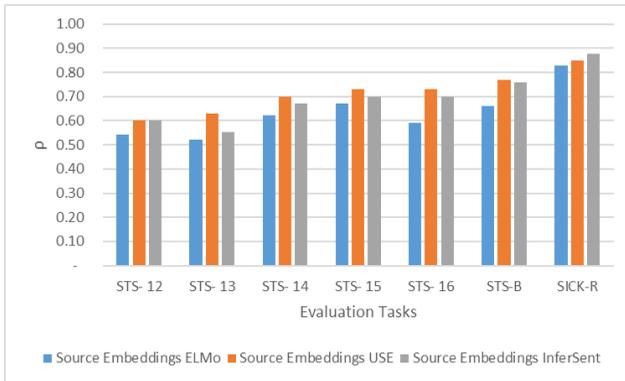

Figure 3: Results of the semantic relatedness and textual similarity tasks for the Source embeddings (Pearson correlation coefficient)

As can be seen in Table 10, where we report the results for the semantic relatedness and textual similarity tasks, the source embedding Universal Sentence Encoder (USE) using Transformer model achieved excellent results on all five STS tasks, except for the SICK-R (semantic relatedness) where InferSent achieved better results. This is in line with the results of [6]. Thus showing that the following results can be used as a benchmark with which the Meta-Embeddings can be compared with.

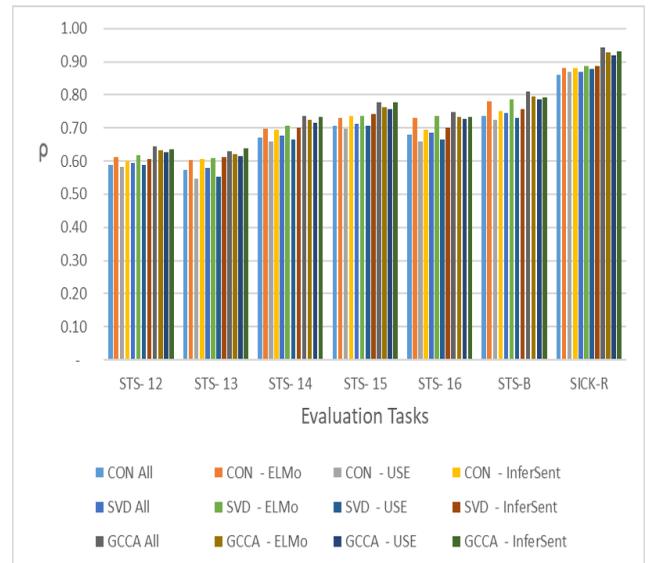

Figure 4: Results of the semantic relatedness and textual similarity tasks for the Ensemble embeddings (Pearson correlation coefficient)

From Table 3 it can be seen CON, the simplest ensemble, has a robust performance. The all-CON embeddings are the weakest variant of CON Embeddings. This is solely due to the curse of dimensionality i.e. 3072+4096+512=7680 sized input which is too many parameters. But the variant ELMo+USE CON performs the best as its size of 3584 seems to be performed efficiently. The ELMo+USE CON outperforms all the three source embeddings in all semantic relatedness and textual similarity tasks.

The SVD beats the CON variant as expected. SVD reduces CON's dimensionality dramatically and still keeps competitive performance, especially on word similarity. SVD maintains a dimension of 3072. The USE+InferSent substantially performs better than all previous variants. As its a combination of two very good representations. It can be seen that the correlation of the other variants is very similar to each other this is because we have reduced all the dimensions. Nonetheless, the process easily beats the benchmark.

From Table 3 it's clear that the GCCA is the best ensemble approach. All four variants perform better than the source embedding. This is due to the fact the GCCA integrates information from data samples that are acquired at multiple feature spaces to produce low-dimensional representations. The best performing GCCA setting was again the ensemble of USE+InferSent. The USE+InferSent GCCA outperforms all the ensembles in all the tasks emerging as a clear winner. Figure 4 gives the comparison of all the Ensemble settings on semantic relatedness and textual similarity tasks.

Moving forward, it can be seen that the dynamic variant, in general, performs better to the ensemble models for the semantic relatedness and textual similarity tasks. Its better performance is due to Bi-LSTM architecture and SNLI training. The DME and CDME variant performs similarly to the GCCA ensemble. Unlike the ensemble models, the total intersection of all the source embeddings performs better to the subset variants. This is because the DME and CDME efficiently capture the information across more dimensions without the issue of dimensionality. The ALL- DME performs best overall in five out of the six textual similarity tasks. And CDME performs the best for the semantic relatedness task. Figure 5 gives the comparison of both the dynamic settings on semantic relatedness and textual similarity tasks. The results of the Natural Language Inference i.e. SICK-E and MRPC dataset results are tabulated in Table 4.

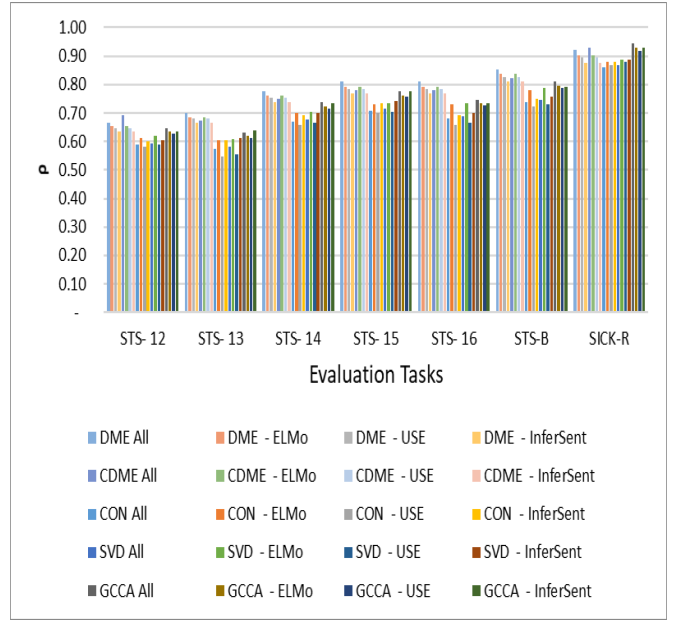

Figure 5: Results of the semantic relatedness and textual similarity tasks for the Dynamic embeddings (Pearson correlation coefficient )

Table 4: Results of the Natural Language Inference tasks (Accuracy )

|  |  | SICK-E | MRPC |
|---|---|---|---|
| **Source Embeddings** | ELMo | **80.16** | **69.33** |
|  | USE | **79.99** | **70.89** |
|  | InferSent | **85.09** | **73.46** |
| **CON** | All | 82.56 | 71.94 |
|  | - ELMo | 84.19 | 73.62 |
|  | - USE | 84.28 | 72.82 |
|  | - InferSent | 84.08 | 73.61 |
| **SVD** | All | 83.38 | 72.65 |
|  | - ELMo | 85.02 | 74.34 |
|  | - USE | 85.10 | 73.54 |
|  | - InferSent | 84.88 | 74.31 |
| **GCCA** | All | **90.74** | **79.06** |
|  | - ELMo | 89.10 | 77.64 |
|  | - USE | 88.29 | 76.93 |
|  | - InferSent | 88.88 | 77.82 |
| **DME** | All | 89.35 | 77.14 |
|  | - ELMo | 87.56 | 75.59 |
|  | - USE | 86.67 | 74.82 |
|  | - InferSent | 84.88 | 73.28 |
| **CDME** | All | **90.20** | **77.87** |
|  | - ELMo | 87.56 | 75.59 |
|  | - USE | 86.67 | 74.82 |
|  | - InferSent | 84.88 | 73.28 |

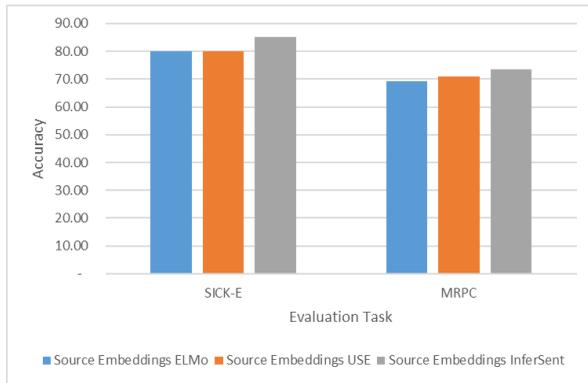

Figure 5: Results of the Natural Language Inference for the Source embeddings (Accuracy )

From Table 4 and Figure 5 it can be seen that the Source embedding InferSent is substantially a better performer than the other two in NLI tasks. Figures 6 & 7 provide a comparison for the NLI task on the ensemble and dynamic Meta-Embedding.

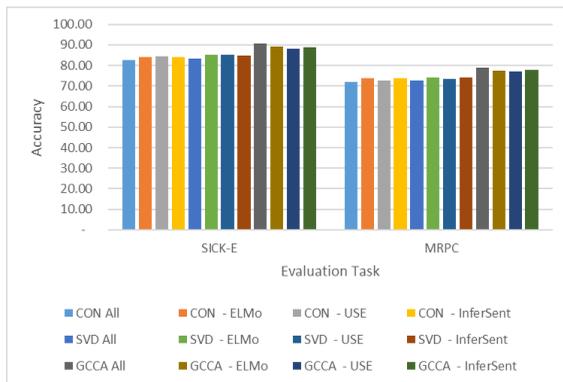

Figure 6: Results of the Natural Language Inference for the Ensemble embeddings (Accuracy ).

For the Ensemble models both SVD and GCC perform better than the SOTA models. This is largely due to the better representation achieved by the dimension reduction techniques. Even a naïve method of CON produces close to SOTA results especially (ELMO+InferSent). The GCCA- All variant is the best ensemble for NLI technique. The GCCA outperforms the dynamic models too in the entailment task displaying the immense power of dimension reduction techniques. For the Dynamic models both the contextualized DME performs better than the DME, unlike the previous task. This shows the need for a self-attention mechanism that is incorporated in the CDME. The All CDME performs better than all the remaining dynamic models.

We can finally conclude that the GCCA method of Ensemble performs better than Concatenation and singular value decomposition for Natural Language Inference and Semantic Similarity tasks especially a complete ensemble. For dynamic ensembles, there is no clear winner as DME performs best on Semantic Similarity tasks while the contextualized upgrade performs better in NLI tasks.

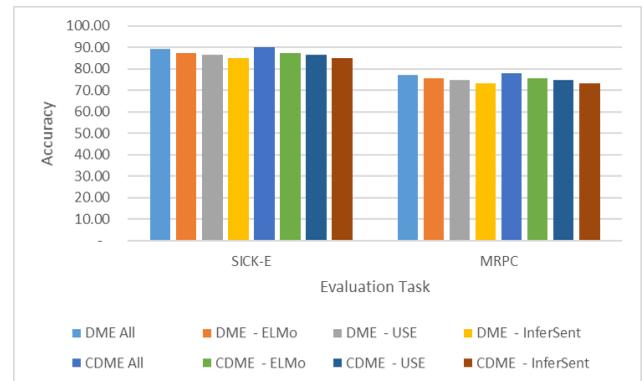

Figure 7: Results of the Natural Language Inference for the Dynamic embeddings (Accuracy )

## 6 CONCLUSIONS:

In this paper, we demonstrated how the use of Meta-Embeddings can better the results of SOTA models on standard NLI tasks. In this paper, we see how the power of multiple individual representations can be aggregated to form one strong representation. For future work, we can utilize several other representations and try to find the upper limit of aggregation possible. Furthermore, we can try to see the effect of Meta-Embeddings various other NLP tasks and try to build a single powerful representation